 \documentclass[accepted]{uai2022} 

\usepackage[american]{babel}

\usepackage{natbib} 
\usepackage{mathtools} 
\usepackage[overload,ntheorem]{empheq}
\usepackage{amsfonts}
\usepackage{amsthm}
\usepackage{algorithm}
\usepackage[noend]{algpseudocode}
\usepackage{subfig}
\usepackage{comment}

\newtheorem{theorem}{Theorem}[section]
\newtheorem{definition}[theorem]{Definition}
\newtheorem{proposition}[theorem]{Proposition}
\newtheorem{example}[theorem]{Example}
\newtheorem{lemma}[theorem]{Lemma}

\newtheorem{corollary}[theorem]{Corollary}

\DeclareMathOperator{\pa}{pa}
\DeclareMathOperator{\ch}{ch}
\DeclareMathOperator{\an}{an}
\DeclareMathOperator{\de}{de}
\DeclareMathOperator{\cum}{cum}
\DeclareMathOperator{\ttop}{top}
\DeclareMathOperator{\argmax}{arg\,max}

\DeclareMathOperator{\var}{var}
\DeclareMathOperator{\rank}{rank}

\usepackage{booktabs} 
\usepackage{tikz} 
\usetikzlibrary{arrows.meta}

\newcommand{\indep}{\perp \!\!\! \perp}
\numberwithin{equation}{section}

\title{Learning Linear Non-Gaussian Polytree Models}

\author[1]{Daniele Tramontano}
\author[2]{Anthea Monod}
\author[1]{\href{mailto:<mathias.drton@tum.edu>?Subject=Your UAI 2022 paper}{Mathias Drton}{}}

\affil[1]{%
Department of Mathematics and Munich Data Science Institute\\
Technical University of Munich\\
Germany
}
\affil[2]{%
Department of Mathematics\\
Imperial College London\\
UK
}

\begin{document}
\maketitle

\begin{abstract}
In the context of graphical causal discovery, we adapt the versatile framework of linear non-Gaussian acyclic models (LiNGAMs) to propose new algorithms to efficiently learn graphs that are polytrees.  Our approach combines the Chow--Liu algorithm, which first learns the undirected tree structure, with novel schemes to orient the edges.  The orientation schemes assess algebraic relations among moments of the data-generating distribution and are computationally inexpensive.
We establish high-dimensional consistency results for our approach and compare different algorithmic versions in numerical experiments.
\end{abstract}

\section{Introduction}\label{sec:intro}


Directed acyclic graphs (DAGs) have been extensively used in causal modeling; the nodes of a graph represent the random variables of the model while the edges represent directed causal effects from source to sink. These causal effects of the parent nodes on the children are quantified by structural equations. 
In this paper, we take up this framework and study the problem of inferring the graphical structure underlying the causal model, given only observational data \citep{drton:maathuis:2017}.  Referred to as structure learning or causal discovery, it is a problem that is difficult due to the statistical curse of dimensionality and computational issues.  Effective methods, thus, need to exploit restrictions on the random variables, graphical structure, or structural equations to simplify the problem 
\citep{Pearl:Primer:2016,Peters:Elements:2017}.  Here, we consider a class of tree-structured graphs, together with linear structural equations where the error terms are mutually independent and non-Gaussian.

Specifically, we work in the versatile causal discovery framework of {\em linear non-Gaussian acyclic models (LiNGAMs)} \citep{shimizu:hoyer:2006,shimizu:2008}.  LiNGAMs postulate linear structural equations with non-Gaussian noise terms to describe the relationships among observed variables.  The non-Gaussianity assumption allows for consistent estimation of the graph encoding the model from observational data alone and for efficient structure learning algorithms \citep[e.g.,][]{shimizu:2011,hyv:smith:2013,wang:drton:2020,hoyer:additive:2008}.  Since the complexity of the structure learning problem depends directly on the underlying graph, consistency results for causal discovery algorithms often require some restrictions on the graph, particularly, when high-dimensional consistency results are desired.  In this context,
the subset of DAGs whose underlying skeleton is a tree---a {\em polytree}---is the most scalable setting, offering low computational complexity whilst retaining model expressiveness \citep{pearl:reasoning:1988}.  In this paper, we propose algorithms to learn a polytree underlying a LiNGAM model.

Learning a polytree may be decomposed into two tasks: extracting the skeleton and determining the orientation of the edges \citep{rebane:pearl:1987,jakobsen:2021}.  Recovering the underlying skeleton may be achieved via the {\em Chow--Liu algorithm} \citep{chow:liu:1968}
.  Existing methods for edge orientation entail checking conditional independence, which is usually carried out by serial hypothesis testing and impacts computational efficiency.  We instead
proceed by exploiting 
recent insights concerning algebraic relations among moments to determine edge orientation  \citep{robeva:2021,amendola:drton:2021,wiedermann:2015,dodge:2001}.  The result is an efficient approach that adapts a classical algorithm to recover the core causal tree structure and augments it with a novel algebraic strategy to determine edge orientation.  The proposed algorithms learn the polytree from observational data alone, in a far more scalable manner than existing LiNGAM algorithms that learn more general graph structures.

The remainder of the paper is organized as follows.  Section~\ref{sec:background} sets the background and theory. Section~\ref{sec:new-algs} presents our contributions where a general population version and three algorithmic scenarios are studied in detail.  Corresponding theoretical guarantees for our proposed algorithms are given in Section~\ref{sec:learning-data}.  Results of numerical experiments are presented in Section~\ref{sec:numerical-experiments}.  We close with a discussion and suggestions for future research in Section~\ref{sec:conclusion}. The proofs of all the results are provided in the Appendix~\ref{app:proofs}, which is part of the supplementary material. Appendix~\ref{app:sec:samp:alg} in the supplementary material gives the detailed description of the sample versions of the algorithms considered in the paper.

\section{Linear Non-Gaussian Structural Causal Models}
\label{sec:background}


A directed graph (digraph) is a pair \(G=(V,E)\), where $V$ is the set of vertices and $E\subset V\times V$ is the set of directed edges.  We let $V=[p]:=\{1,\dots,p\}$. An element $(i,j)\in E$ may also be denoted by $i\xrightarrow{}j$.  A digraph $G$ is acyclic (i.e., a DAG) if it does not contain any directed cycle: there is no sequence of vertices $i_0,\dots,i_k$ with $i_j\xrightarrow{}i_{j+1}\in E$ for $j=0,\dots,k-1$ and $i_0=i_k$.  A path in $G$ is a sequence of vertices $i_0,\dots,i_k$ such that $i_j\xrightarrow{}i_{j+1}\in E$ or $i_{j+1}\xrightarrow{}i_{j}\in E$ for all $j$.  It is directed if all the arrows point in the same direction.  A \emph{polytree} is a DAG in which there is a unique path between any two vertices.  

If $i\xrightarrow{} j\in E$, then $i$ is a parent of $j$, and $j$ is a child of $i$. If $G$ contains a directed path from $i$ to $j$, then $i$ is an ancestor of $j$ and $j$ is a descendant of $i$. The sets of parents, children, ancestors, and descendants of $i\in V$ are denoted by $\pa(i), \ch(i), \an(i), \de(i)$, respectively.

Let $X=(X_i)_{i\in[p]}$ be a random vector indexed by the vertices of a DAG $G$. For $A\subset [p]$, let $X_A=(X_i)_{i\in A}$.  
When $X_A$ is conditionally independent of $X_B$ given $X_C$ for disjoint subsets $A,B,C\subset [p]$, we write $A\indep B|\,C$.
The joint distribution of $X$ 
satisfies the local Markov property with respect to $G$ if
$
    \{i\} \indep [p]\setminus(\pa(i)\cup \de(i))\ |\ \pa(i)\ \forall\ i\in[p].
$
The Markov equivalence class of $G$ is the set of all DAGs that encode the same conditional independence relations, i.e., for which the set of distributions satisfying the local Markov property is the same.  See \citet[Chap.~1]{handbook} for further details.

The skeleton of a DAG is the undirected graph obtained by replacing each directed edge by an undirected edge.  Here, edges are denoted by $\{i,j\}\subseteq E$.

\subsection{Structural Equations}
\label{subsec:equations}


A structural equation model hypothesizes that every random variable in $X$ is functionally related to its parent variables:
$
    X_i=f_i(X_{\pa(i)},\varepsilon_i), \ i\in V,
$
where the $\varepsilon_i$ are independent noise terms and the $f_i$ are measurable functions.  If the $f_i$ are linear, then we obtain a linear structural equation model (LSEM). An LSEM can be written in matrix form as
\begin{equation}
    \label{eq:lsem}
    X=(I-\Lambda)^{-\top}\varepsilon,
\end{equation}
where $\Lambda=(\lambda_{ij})$ with $\lambda_{ij}\neq0$ only if $i\to j\in E$.
An LSEM constrains the dependence structure on the coordinates of $X$, but not the mean.  Hence, when working with the LSEM, we may assume without loss of generality that $\mathbb{E}[\varepsilon_i]=0$, which implies $\mathbb{E}[X_i]=0$ for all $i\in V$. 

Let $\varepsilon^{(2)}=(\mathbb{E}[\varepsilon_i\varepsilon_j])_{ij}$ be the covariance matrix of $\varepsilon$, which is a diagonal matrix by independence, and write $\varepsilon^{(2)}_i:=\mathbb{E}[\varepsilon_i^2]>0$ for its $i$th diagonal entry.  The covariance matrix of $X$ is then the positive definite matrix
\begin{equation}
\label{eq:Sigma}
    \Sigma=(I-\Lambda)^{-\top}\varepsilon^{(2)}(I-\Lambda)^{-1}.
\end{equation}



\subsection{Cumulants in Gaussian and Non-Gaussian Models}
\label{subsec:non-gaussian}



Cumulants are alternative representations of moments of a distribution.  Here, we formalize the definition in higher order settings and discuss their implications under Gaussian and non-Gaussian errors.

\begin{definition}
The $k$th cumulant tensor of a random vector $(X_1,\dots,X_p)$ is the $k$-way tensor in $\mathbb{R}^{p\times\dots\times p}\equiv(\mathbb{R}^p)^k$ whose entry 
in position $(i_1,\dots,i_k)$ is the joint cumulant
\begin{equation*}
    \begin{aligned}
           &\cum(X_{i_1},\dots,X_{i_k}):=\\
           &\sum_{(A_1,\dots,A_L)}(-1)^{L-1}(L-1)!\mathbb{E}\bigg[\prod_{j\in A_1} X_j\bigg]\cdots\mathbb{E}\bigg[\prod_{j\in A_L} X_j\bigg],
    \end{aligned}
\end{equation*}
where the sum is taken over all partitions $(A_1,\dots, A_L)$ of the multiset $\{i_1,\dots,i_k\}$.
\end{definition}

In our context, the variables have mean $0$, so 
\begin{align*}
   \cum(X_i) & =\mathbb{E}[X_i]=0,\\
   \cum(X_{i_1},X_{i_2}) &=\mathrm{Cov}[X_{i_1},X_{i_2}]=\mathbb{E}[X_{i_1}X_{i_2}].
\end{align*}
More generally, the sum can be restricted to the partitions in which all blocks $A_i$ have at least two elements.  In particular, 
\begin{equation*}
    \begin{aligned}
        &\cum(X_{i_1},X_{i_2},X_{i_3}) =\mathbb{E}[X_{i_1}X_{i_2}X_{i_3}],\\
        &\cum(X_{i_1},X_{i_2},X_{i_3},X_{i_4}) =\mathbb{E}[X_{i_1}X_{i_2}X_{i_3}X_{i_4}]\\
        &-\mathbb{E}[X_{i_1}X_{i_2}]\mathbb{E}[X_{i_3}X_{i_4}]-\mathbb{E}[X_{i_1}X_{i_3}]\mathbb{E}[X_{i_2}X_{i_4}]\\
        &-\mathbb{E}[X_{i_1}X_{i_4}]\mathbb{E}[X_{i_2}X_{i_3}].
    \end{aligned}
\end{equation*}

The following powerful result dictates a simple condition for Gaussianity of $X$.

\begin{theorem}{\cite[Theorem 2]{marcinkiewicz:1939}}
\label{Marcinkiewicz}
If there exists $k$ such that $\cum(X_{i_1},..,X_{i_j})=0$ for all $j\geq k$, then $k=3$ and $X$ has a multivariate Gaussian distribution.
\end{theorem}

Furthermore, the following results dictate when the assumptions of Theorem \ref{Marcinkiewicz} are satisfied, thus giving rise to Gaussianity, especially under LSEMs.

\begin{lemma}
\label{lem:indep_cum}
If the variables $ \varepsilon_1,\dots,\varepsilon_n$ are independent, then $\cum(\varepsilon_{i_1},\dots,\varepsilon_{i_k})=0$ unless $i_1=\dots=i_k$.
\end{lemma}

\begin{lemma}
\label{lem:tucker}
Let the random vector $X$ follow the LSEM from \eqref{eq:lsem} with noise vector $\varepsilon$.  Let 
$\mathcal{C}^{(k)}$ and $\varepsilon^{(k)}$ be the $k$th order cumulant tensors of $X$ and $\varepsilon$, respectively.  Then
\begin{align*}
    \mathcal{C}^{(k)}&= \varepsilon^{(k)}\bullet \big[(I-\Lambda)^{-1} \big]_{j=1}^k\\
    &=\varepsilon^{(k)}\bullet(I-\Lambda)^{-1}\bullet \dots \bullet (I-\Lambda)^{-1}
\end{align*}
is the Tucker product of $\varepsilon^{(k)}$ and $k$ copies of $(I-\Lambda)^{-1}$.
\end{lemma}
Notice here that $\mathcal{C}^{(k)}$ reduces to \eqref{eq:Sigma} when $k=2$.

See \citet{comon:jutten:handbook} and references therein for proofs of Theorem \ref{Marcinkiewicz} and Lemmas \ref{lem:indep_cum} and \ref{lem:tucker}.

The next definition introduces the cumulant model obtained from the LSEM \eqref{eq:lsem}.  

\begin{definition}
Let $G=(V,E)$ be a DAG, and let $K\geq2$ be an integer.  The $K$th cumulant model of $G$ is the set of $K$-way tensors 
\begin{multline*}
    \mathcal{M}^{(K)}(G)=
    \{\varepsilon^{(K)}\bullet \big[(I-\Lambda)^{-1} \big]_{j=1}^K\;:\;\\
    \Lambda\in\mathbb{R}^E,\; \varepsilon^{(K)}\in(\mathbb{R}^{p})^K \ \text{diagonal}\}.
\end{multline*}
Here, $\mathbb{R}^E$ is the set of $p\times p$ matrices with support $E$.
Further, the cumulants up to order K defined by G are modeled by 
\begin{equation}
    \mathcal{M}^{(\leq K)}(G)=\mathcal{M}^{(2)}(G)\times\dots\times\mathcal{M}^{(K)}(G).
\end{equation}
\end{definition}

By Theorem~\ref{Marcinkiewicz}, all multivariate Gaussian vectors $X$ correspond to the zero element of $\mathcal{M}^{(K)}(G)$ for $k\geq3$.  

When the errors in an LSEM are Gaussian, all distributional information is captured by the covariance matrix and equivalence issues arise that hinder identifiability of the full graph.  
It then becomes necessary to consider non-Gaussian settings.  Relaxing the constraint of Gaussianity gives rise to the class of LiNGAMs where the underlying graph now becomes identifiable \citep{shimizu:hoyer:2006,shimizu:2011}.  We will exploit this property algorithmically and use the signal provided by higher cumulants; we do this by way of {\em treks}.

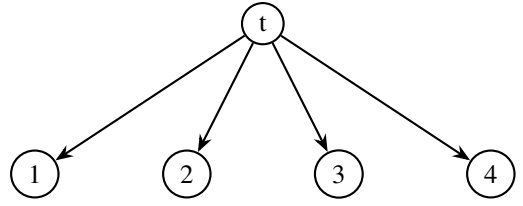
\begin{figure}[ht]
\centering
\begin{tikzpicture}[scale=0.5]
\begin{scope}[every node/.style={circle,thick,draw}]
    \node (1) at (0,0) {t};
    \node (2) at (-6,-4) {1};
    \node (3) at (-2,-4) {2};
    \node (4) at (2,-4) {3};
    \node (5) at (6,-4) {4};
\end{scope}

\begin{scope}[>={Stealth[black]},
              every edge/.style={draw=black,thick}]
    \path [->] (1) edge node{} (2);
    \path [->] (1) edge node{} (3);
    \path [->] (1) edge node{} (4);
    \path [->] (1) edge node{} (5);
\end{scope}
\end{tikzpicture}
\caption{Example of a 4-trek.}
\label{fig:trek}
\end{figure}

\begin{definition}[Multi-Trek]
A $k$-trek between vertices $i_1,\dots,i_k\in V$ of a DAG $G=(V,E)$ is a collection of directed paths $T=(P_1,\dots,P_k)$ in $G$ that share the same source and have $i_j$ as the sink of $P_j$ for all $j$. The common source node is the top of the trek $\ttop(T)$.  A trek is simple if the top node is the unique node on all the paths. 
\end{definition}
We denote the set of $k$-treks between $i_1,\dots,i_k$ by $\mathcal{T}(i_1,\dots,i_k)$ and the set of simple treks by $\mathcal{S}(i_1,\dots,i_k)$. See Figure~\ref{fig:trek} for an example.

If $P$ is a directed path in the DAG $G=(V,E)$ and $\Lambda=(\lambda_{ij})\in\mathbb{R}^E$, then $\lambda^P=\prod_{(i,j)\in P}\lambda_{ij}$ is a path monomial. 
For a $k$-trek $T=(P_1,\dots,P_k)$, set $\lambda^T:=\lambda^{P_1}\cdots\lambda^{P_k}$. 


\begin{proposition}[Multi-Trek Rule]
\label{prop:multi:trek}
The $k$th order cumulant tensor $\mathcal{C}^{(k)}(G)$ of $X$ can be expressed as
\begin{equation}
\label{eq:trek}
    \mathcal{C}^{(k)}_{i_1,\dots,i_k}(G)=\sum\varepsilon^{(k)}_{\ttop(T)}\lambda^T,
\end{equation}
where the sum is over all the treks $T$ in $\mathcal{T}(i_1,\dots,i_k)$ and $\varepsilon^{(k)}_{\ttop(T)}$ denotes the $\ttop(T)$ diagonal entry of $\varepsilon^{(k)}$.
\end{proposition}

Proposition \ref{prop:multi:trek} follows from Lemma~\ref{lem:tucker} and expanding the entries of $(I-\Lambda)^{-1}$ into sums of path monomials as in the usual trek rule for covariances \citep{robeva:2021}.

\begin{corollary}[Simple Multi-Trek Rule]
\label{cor:simple-trek-rule}
The $k$th order cumulant tensor $\mathcal{C}^{(k)}(G)$ of $X$ can be expressed as
\begin{equation}
    \mathcal{C}^{(k)}_{i_1,\dots,i_k}(G)=\sum \mathcal{C}^{(k)}_{\ttop(S)}(G)\lambda^{S},
\end{equation}
where the sum is extended to all the simple treks $S$ in $\mathcal{S}(i_1,\dots,i_k)$.
\end{corollary}

\begin{corollary}
\label{cor:simple_trek_2}
The $i$th diagonal entry of $\mathcal{C}^{(k)}$ is
\begin{equation*}
     \mathcal{C}^{(k)}_{i}(G)=\displaystyle\sum_{p_1,\dots,p_k\in \pa(i)}\lambda_{p_1, i}\cdots\lambda_{p_k,i}\mathcal{C}^{(k)}_{p_1,\dots,p_k}(G)+\varepsilon^{(k)}_i.
\end{equation*}
\end{corollary}




\subsection{Polytree Models}
\label{subsec:polytrees}

For general graphs, the algebraic relations among the cumulants may be far more complicated than the bivariate case (which is discussed in Example~\ref{ex:two:vert}) and have not yet been fully characterized.  However, there exists a generalization of rank-one constraints for polytrees, which we now discuss.  


By consequence of there being at most one directed path between any two nodes of a polytree $G$, there is at most one simple trek between any set of nodes $i_1,\dots,i_k$.  The simple multi-trek rule then reduces to $C^{(k)}_{i_1,\dots,i_k}(G)=\lambda^{S}\mathcal{C}^{(k)}_{\ttop(S)}$ for a trek between nodes with $S$ being the unique simple trek; denote the top of the simple trek between $i_1,\dots,i_k$, if it exists by $\ttop(i_1,\dots,i_k)$.  Also, $C^{(k)}_{i_1,\dots,i_k}(G)=0$ if there is no $k$-trek between the nodes.

For any two  vertices $i\not=j$, let $c^{(i,j),k}_m$ denote the $k$th order cumulant $\mathcal{C}^{(k)}_{{i\dots i},{j\dots j}}(G)$, where the first $m$ indices are equal to $i$ and the remaining $m-k$ equal $j$.   

\begin{proposition}
\label{prop:rank}
Let $e: i\to j$ be an edge of a polytree $G$.  Then the following matrix is of rank one
\begin{align}
\label{matrix}
    A^{e,K}=\left[
\begin{matrix}
c^{e,k}_m \\
c^{e,k}_{m-1}
\end{matrix} \mid 2\leq m\leq k\leq K 
\right].
\end{align}
\end{proposition}

The first column of $A^{e,K}$ contains $\mathbb{E}[X_i^2]>0$.  Moreover, for every distribution induced by non-Gaussian errors, there exists $k$ such that 
$\mathcal{C}^{(k)}_i\neq0$.  Hence, at least one minor of $A^{e,K}$ gives us an equation that is satisfied if $i\to j$ is in $G$, and is not satisfied in general for the graph with the edge reversed.  This observation will provide the foundation for our learning algorithm, which we now present. 

\section{Learning Non-Gaussian Polytrees from Moments}
\label{sec:new-algs}


We now present our population algorithm for learning polytrees with three versions for learning the edge orientations.  The first common phase is skeleton recovery.

\subsection{Learning the Skeleton}
\label{subsec:chow-liu}

In its original formulation, the Chow--Liu algorithm gives the maximum likelihood tree approximation of a given discrete distribution \citep{chow:liu:1968}.  The tree obtained is the maximum weight spanning tree of the complete undirected graph with edge weights $w(i,j)$, given by the mutual information between $X_i$ and $X_j$.  Under a non-degeneracy assumption, the same Chow--Liu algorithm can be used to recover skeletons in the polytree setting \cite[Theorem 1]{rebane:pearl:1987}; the proof is based on the following property of the mutual information.
\begin{proposition}
If the polytree that defines the model contains the subgraph $i\to j\to l$ or $i\xleftarrow{}j\to l$, then
\begin{equation*}
    \min\{I(X_i,X_j),I(X_j,X_l)\}>I(X_i,X_l),
\end{equation*}
where $I(\cdot,\cdot)$ is the mutual information.
\end{proposition}
When working with an LSEM, a stronger result justifies the use of the absolute value of the correlation coefficient instead of the mutual information.  
\begin{lemma}[Wright's Formula, \citep{wright:1960}]
\label{lem:wright}
In the LSEM defined by a polytree, the correlation $\rho_{i,j}=\text{Corr}[X_i,X_j]$ satisfies
\begin{equation}
    |\rho_{i,j}|=\begin{cases}
        \displaystyle\prod |\rho_e|,\quad &\mathcal{T}(i,j)\neq\emptyset,\\
        0,\quad &\text{otherwise},
        \end{cases}
\end{equation}
where the product is taken over the edges of the unique trek from $i$ to $j$, and $\rho_e$ denotes the correlation between the random variables indexed by the endpoints of the edge $e$.
\end{lemma}
\begin{definition}
Let $R=(\rho_{i,j})$ be the correlation matrix of a random vector $X$.  The Chow--Liu tree $\mathcal{M}(R)$ is the (undirected) maximum weight spanning tree over $[p]$, with weights given by $|\rho_{i,j}|$. 
\end{definition}
Kruskal's algorithm may be applied to compute the Chow--Liu tree \citep{kruskal:1956}.

\begin{proposition}
\label{prop:cl:corr}
Let $R=(\rho_{i,j})$ be the correlation matrix of a random vector $X=(X_1,\dots,X_p)$ that follows the LSEM given by a polytree $G$. If $0<|\rho_{i,j}|<1$ for every $e:i\to j\in E$, then $\mathcal{M}(R)$ equals the  skeleton of $G$.
\end{proposition}

The assumption $|\rho_{i,j}|<1$ holds for all random vectors with positive definite covariance matrix.  Moreover, in a polytree model, $|\rho_{i,j}|>0$ for an edge $(i,j)$ if  $\lambda_{ij}\not=0$.

\subsection{Learning Orientations}
\label{subsec:learning-orientations}


We now present three ways to orient the edges in the estimated skeleton.  The three resulting orientation algorithms are based on Proposition~\ref{prop:rank} and the following result.
\begin{theorem}
\label{theo:generic_cum}
Consider the LSEM given the polytree $G$, and let $e:i\to j$ be an edge of $G$.  Then
\begin{enumerate}
    \item[(i)] $\rank(A^{i\to j,K})=1$,
    \item[(ii)] $\rank(A^{j\to i,K})=2$, for generic edge coefficients and error cumulants up to order $K$.
\end{enumerate}

\end{theorem}
\begin{proposition}
\label{prop:MEC}
 Suppose the skeleton of the polytree $G$ contains the subgraph $i-j-l$ with $\rho_{i,j},\rho_{j,l}\neq0$.  Then the corresponding subgraph of $G$ is $i\to j\xleftarrow{}l$ iff $\rho_{i,l}=0$.

\end{proposition}

We now present \textit{PairwiseOrientation\_Pop};  Algorithm~\ref{alg:pop:pair}.  This algorithm takes as input the list of unoriented edges and the parameter $K\geq3$, which defines the highest order cumulant used in $A^{i\to j,K}$.  It orients each edge separately by checking whether the rank of $A^{i\to j,K}$ is $1$ or not.

\begin{algorithm}\caption{PairwiseOrientation\_Pop$(E,K)$}
\label{alg:pop:pair}
    \begin{algorithmic}[1]
        \State{$O\gets\emptyset$}
        \For{$\{i,j\}\in E$} 
            \If{$\rank(A^{i\to j,K})=1$}
                \State{$O\gets O\cup\{i\to j\}$}
                \Else\,  \State{$O\gets O\cup\{j\to i\}$}
            \EndIf
        \EndFor
    \Return $O$
    \end{algorithmic}
\end{algorithm}


Our second algorithm \textit{TPO\_Pop}, Algorithm~\ref{alg:pop:pair_trip}, proceeds recursively. At each step, it takes the order $K$, a list of already oriented edges $O$, a list of still unoriented edges $E$, and, possibly, an oriented edge $o$, as inputs. Here $t(o)$ is the target/sink of the edge and $E\cap t(o)$ is the (possibly empty) set of unoriented edges containing $t(o)$. The procedure checks if there are unoriented edges, and if so, it searches for triplets of the form $i\to j-k$, where the oriented edge $o=i\to j$ can come either from the previous call of the procedure or from checking the rank of $A^{i\to j,K}$.  For such a triplet, the method determines whether $\rho_{i,k}=0$, orienting the other edge according to the result. The algorithm is initialized with $O=o=\emptyset$ and the full list of undirected edges, $E$.

\begin{algorithm}\caption{TPO\_Pop$(E,K,O,o)$}
\label{alg:pop:pair_trip}
    \begin{algorithmic}[1]
        \If{$E\neq\emptyset$}
            \If{$o=\emptyset$}
                \State{$\{i,j\}\gets E[1]$}
                \If{$\rank(A^{i\to j,K})=1$}
                    \State{$o\gets (i\to j)$}
                    \State{$O\gets O\cup\{o\}$}
                    \Else  
                        \State{$o\gets (j\to i$)}
                        \State{$O\gets O\cup\{o\}$}
                \EndIf
            \EndIf
            \State{$E_o\gets E\cap t(o)$}
            \If{$E_o\neq\emptyset$}
                \State{$E\gets E\setminus E_o$}
                \For{$t(o)-k\in E_o$}
                    \If{$\rho_{s(o),k}=0$}
                        \State{$O\gets O\cup\{k\to t(o)\}$}
                        \Else                                                       \State{$O\gets O\cup\{t(o)\to w\}$}
                            \State{$o\gets(t(o)\to w)$}
                            \State{$O,E\gets TPO\_Pop(E,K,O,o)$}
                    \EndIf
                \EndFor
            \EndIf
            \State{$O,E\gets TPO\_Pop(E,K,O,\emptyset)$}
        \EndIf
        \Return{O,E}
    \end{algorithmic}
\end{algorithm}

Our third proposed algorithm \textit{PTO\_Pop}, Algorithm~\ref{alg:pop:trip_pair}, can be seen as a direct extension of learning completed partially directed graphs (CPDAG)---a mixed graph that encodes the causal information common to all the members of a Markov equivalence class.  Here, we first compute the CPDAG following \cite{rebane:pearl:1987}, then we orient all remaining undirected edges by considering the rank of $A^{i\to j,K}$.  This ensures that no other unshielded colliders appear.

\begin{algorithm}\caption{PTO\_Pop$(E,K)$}
\label{alg:pop:trip_pair}
    \begin{algorithmic}[1]
        \State{$O\gets\emptyset$}
        \For{$i-j-k\in E$}
            \If{$\rho_{i,k}=0$}
                \State{$E\gets E\setminus\{\{i,j\},\{j,k\}\}$}
                \State{$O\gets O\cup\{i\to j, k\to j\}$}
            \EndIf
        \EndFor
        \For{$i\to j\in O$}
            \For{$j-l\in E$}
                \State{$E\gets E\setminus\{(j,l)\}$}
                \State{$O\gets O\setminus\{j\to l\}$}
            \EndFor
        \EndFor
        \For{$\{i,j\}\in E$}
            \If{$\rank(A^{i\to j,K})=1$}
                \State{$O\gets O\cup\{i\to j\}$}
                \For{$j-l\in E$}
                    \State{$E\gets E\setminus\{(j,l)\}$}
                    \State{$O\gets O\setminus\{j\to l\}$}
                \EndFor
                \Else 
                    \State{$O\gets O\cup\{j\to i\}$}
                    \For{$i-l\in E$}
                        \State{$E\gets E\setminus\{(i,l)\}$}
                        \State{$O\gets O\setminus\{i\to l\}$}
                    \EndFor
            \EndIf
        \EndFor
        \Return{O}
    \end{algorithmic}
\end{algorithm}

The following example compares our three algorithms.

\begin{example}
Consider the graph $G$ with  $1\xrightarrow{}2\xrightarrow{}3\xleftarrow{}4$.  With the skeleton $1-2-3-4$ inferred, the algorithm \textit{PairwiseOrientation\_Pop} sequentially computes the rank of $A^{i\to j,K}$ in the order of all edges and orients them according to the results.  \textit{TPO\_Pop} orients $1-2$ using the rank condition and then checks if $\rho_{1,3}=0$.  Since this is not the case, it orients $2-3$ using the rank condition and then $3-4$ checking that $\rho_{2,4}=0$.  Finally, 
\textit{PTO\_Pop} first computes $\rho_{1,3}$ and $\rho_{2,4}$.  Since $\rho_{2,4}=0$, it orients $2-3-4$, and then orients $1-2$ with the rank condition.
\end{example}
\begin{theorem}
The three versions of the algorithm are correct for generic edge coefficients and cumulants up to order $K$.
\end{theorem}

\section{Learning Non-Gaussian Polytrees from Data}
\label{sec:learning-data}


We now consider the empirical versions of our algorithms, which now learn a polytree from a dataset consisting of $n$ i.i.d.~random vectors.  The algorithms then process the sample correlations $\hat\rho_{i,j}$ and sample cumulants $\hat c^{(i,j),k}_m$.  Let $\hat\Sigma_{i,j}$ be the unbiased sample covariances.  Then $\hat\rho_{i,j}=\hat\Sigma_{i,j}/\sqrt{\hat\Sigma_{i,i}\hat\Sigma_{j,j}}$.  Generalizing sample covariances, we take the sample cumulants $\hat c^{(i,j),k}_m$ to be the $k$-statistics that estimate $c^{(i,j),k}_m$ in an unbiased manner \citep[\S4.2]{mccullagh:tensors}.

We provide consistency results in a high dimensional setting where the size of the polytree grows at a faster rate than the sample size, subject to log-concavity of the variables.  Specifically, we assume the errors $\varepsilon_i$ and thus also the observation vector $X$ are log-concave distributed.  This setting allows for the following corollary that builds on the concentration inequality given in Lemma B.3 of \cite{lin:drton:2016}. 

\begin{corollary}
\label{cor:log:cum}
Let $K\in\mathbb{N}$ and suppose that all moments up to order $2K$ of the random vector $X$ are bounded in magnitude by a constant $M_K>0$. There exists a constant $L>0$ such that for any $k\le K$, if $\hat{c}^{(i,j),k}_m=c^{(i,j),k}_m+\epsilon^{(i,j),k}_m$ is the  $k$-statistic for a sample of size $n$, for every $\delta>0$ where
$\displaystyle
    \frac{2}{LK^2\sqrt{M_K}}\left(\frac{\delta\sqrt{n}}{e}\right)^{\frac{1}{K}}>2,
$
we have
\begin{equation*}
    \mathbb{P}[|\epsilon^{(i,j),k}_m|>\delta]\leq\exp\left\{-\frac{2}{LK^2\sqrt{M_K}}\left(\delta\sqrt{n}\right)^{\frac{1}{K}}\right\}.
\end{equation*}
\end{corollary}


\subsection{Learning the Skeleton Consistently}

Let $\rho_{\min}$ and $\rho_{\max}$ be the respective minimum and maximum of the absolute edge correlations in the set $\{|\rho_{i,j}|:i\to j\in E\}$ with $0<\rho_{\min},\rho_{\max}<1$. We will use the following lemma on the correctness of the Chow--Liu tree $\mathcal{M}(R_n)$ computed from the sample correlation matrix $R_n=(\hat\rho_{i,j})$, together with Lemma 7 from \cite{harris:drton:2013}, both restated below.
\begin{lemma}
\label{lemma:lower:bound}
Let $\gamma=\rho_{\min}(1-\rho_{\max})/2$.  Then the event $F:=\bigcap\{|\hat{\rho}_{i,j}-\rho_{i,j}|\le\gamma\}$ satisfies $F\subset \{\mathcal{M}(\hat R_n)=\mathcal{S}(G)\}$.
\end{lemma}
\begin{lemma}
\label{lemma:sym:matrices}
If $A,B$ are $2\times2$ symmetric matrices, with $A$ positive definite, $a_{1,1},a_{2,2}\ge 1$, and $||A-B||_{\infty}<\delta$, then
\begin{equation}
    \left|\frac{a_{1,2}}{\sqrt{a_{1,1}a_{2,2}}}-\frac{b_{1,2}}{\sqrt{b_{1,1}b_{2,2}}}\right|<\frac{2\delta}{1-\delta}.
\end{equation}
\end{lemma}

We now have the following consistency result for the Chow--Liu tree $\mathcal{M}(R_n)$.

\begin{theorem}
\label{theo:chow:cons}
Let $\lambda:=\min\{\min_i\Sigma_{i,i},1\}$ and let $\gamma$ and $M_2$ be defined as in Lemma~\ref{lemma:lower:bound} and Corollary~\ref{cor:log:cum} respectively.  Then
\begin{equation*}  
    \begin{aligned}
        &\mathbb{P}(\mathcal{M}(R_n)=\mathcal{S}(G))\\
        &\geq1-\frac{3p(p-1)}{2}\exp\left\{-\frac{1}{2L\sqrt{M_2}}\left(\frac{\lambda\gamma\sqrt{n}}{2+\lambda}\right)^{\frac{1}{2}}\right\},
    \end{aligned}
\end{equation*}
for all 
$
    n> \frac{e^2(2+\lambda)^2(4L^2\sqrt{M_2})^4}{\lambda^2\gamma^2}.
$

\end{theorem}

\subsection{Learning Orientations Consistently}

For every edge $e=\{i,j\}$ in the skeleton $\mathcal{S}(G)$, let $v_r(e), v_w(e)\in\mathbb{R}^{B(K)}$ be the vectors containing the minors of $A^{(r(e)),K}$ and $A^{(w(e)),K}$ involving the first column.  Here, $r(e)$ and $w(e)$ are the correct and incorrect orientations of $e$ in $G$, respectively.  Let $B(K)=K(K-1)/2-1$ be the size of the vectors. 

We assume that there exists $\delta>0$ such that $||v_w(e)||>\delta$ for all $e\in \mathcal{S}(G)$, where $||\cdot||$ is the 2-norm. Let $M_K$, $L$, and $\epsilon^{(i,j),k}_m$ be defined as in  Corollary~\ref{cor:log:cum}. Moreover, let $c$ be the vector containing all the cumulants $c^{(i,j),k}_m$ such that edge $\{i,j\}\in\mathcal{S}(G)$ and $0\leq m\leq k\leq K$. Write $\hat{c}_n$ for the vector containing the sample versions of these cumulants.  Finally, let $\epsilon_n$ be the corresponding error vector tracking the differences between the true and sample cumulants.

\begin{lemma}
\label{lem:taylor}
If $f$ is the difference of two monomials of degree $2$ in the variables $c$, then
\begin{equation}
    \left|f(c+\epsilon_n)-f(c)\right|\leq 4M_K||\epsilon_n||_{\infty}+2||\epsilon_n||_{\infty}^2.
\end{equation}

\end{lemma}

For use with data, the proposed algorithms in Section \ref{sec:new-algs} must be modified to allow for sampling variability. In particular, instead of assessing whether or not $\rank(A^{i\to j,K})=1$, we check  $||\hat{v}_{i\to j}(\{i,j\})||<||\hat{v}_{j\to i}(\{i,j\})||$ instead.  Here, $\hat v$ is the sample analogue of $v$, computed using sample moments.  Similarly, for the independence test (vanishing of correlation), we check whether or not the absolute sample correlation is below a threshold $\rho_{\theta}$;  Lemma~\ref{lem:rho_crit} clarifies the possible choices of the threshold. The resulting sample versions of the algorithms are given in Appendix~\ref{app:sec:samp:alg}.

Let $\mathcal{A}_n^{PO}(E,K)$ be the output of Algorithm~\ref{alg:samp:pair} applied to a sample of size $n$ and let $E_{\mathcal{S}(G)}$ be the edge set of the true skeleton of $G$.  Then we have the following consistency result.

\begin{lemma}
\label{lem:po:lower:bound}
Let $\delta':=\min\{\frac{\delta}{4M_K\sqrt{B(K)}},\frac{\sqrt{\delta}}{\sqrt[4]{4B(K)}}\}$. Then
\begin{equation*}
    \begin{aligned}
    &\mathbb{P}(\mathcal{A}_n^{PO}(E_{\mathcal{S}(G)},K)=G)\\
    &\geq 1-4B(K)(p-1)\exp\left\{-\frac{2}{LK^2\sqrt{M_K}}\left(\delta'\sqrt{n}\right)^{\frac{1}{K}}\right\},
    \end{aligned}
\end{equation*}
for all 
$
     n>   \frac{e^2(LK^2\sqrt{M_K})^{2K}}{\delta^{'^2}}.
$
\end{lemma}

\begin{theorem}
\label{theo:po:cons}
Suppose the data are an $n$-sample drawn from a distribution in the LSEM given by a polytree $G$.  Let $\hat G$ be the polytree obtained by applying Algorithm~\ref{alg:samp:pair} to the (undirected) edge set of the Chow--Liu tree $\mathcal{M}(R_n)$.  Then $\hat G=G$ with probability greater than 
\begin{equation*}
    \begin{aligned}
         &1-4B(K)(p-1)\exp\left\{-\frac{2}{LK^2\sqrt{M_K}}\left(\delta'\sqrt{n}\right)^{\frac{1}{K}}\right\}\\
         &-\frac{3p(p-1)}{2}\exp\left\{-\frac{1}{2L\sqrt{M_2}}\left(\frac{\lambda\gamma\sqrt{n}}{2+\lambda}\right)^{\frac{1}{2}}\right\},
    \end{aligned}
\end{equation*}
for all 
$
        n> \max\left\{{\tfrac{e^2(2+\lambda)^2(4L^2\sqrt{M_2})^4}{\lambda^2\gamma^2}, \tfrac{e^2(LK^2\sqrt{M_K})^{2K}}{\delta^{'^2}}}\right\},
$
with constants defined in Lemma~\ref{lem:po:lower:bound} and Theorem~\ref{theo:chow:cons}.
\end{theorem}

\begin{lemma}
\label{lem:rho_crit}
Let $\tilde{\gamma}=\min\{\rho_{\min}/3,(1-\rho_{\max})/2\}\rho_{\min}$, and let $\lambda$ and $M_2$ be as in Theorem~\ref{theo:chow:cons}. If $\tilde{\gamma}<\rho_{\theta}<\rho_{\min}^2-\tilde{\gamma}$,  then the probability that all independence tests carried out by Algorithm~\ref{alg:samp:trip_pair} yield correct decisions is bounded from below by
\begin{equation*}
    1-\frac{3p(p-1)}{2}\exp\left\{-\frac{1}{2L\sqrt{M_2}}\left(\frac{\lambda\Tilde{\gamma}\sqrt{n}}{2+\lambda}\right)^{\frac{1}{2}}\right\},
\end{equation*}
for all
$
         n> \frac{e^2(2+\lambda)^2(4L^2\sqrt{M_2})^4}{\lambda^2\tilde{\gamma}^2}.
$
The same statement holds for Algorithm~\ref{alg:samp:pair_trip}.
\end{lemma}

\begin{theorem}
\label{theo:pt_tp:cons}
Suppose the data are an $n$-sample drawn from a distribution in the LSEM given by a polytree $G$.  Let $\hat G$ be the polytree obtained by applying Algorithm~\ref{alg:samp:trip_pair} or~\ref{alg:samp:pair_trip} to the (undirected) edge set of the Chow--Liu tree $\mathcal{M}(R_n)$.  If the threshold satisfies $\tilde{\gamma}<\rho_{\theta}<\rho_{\min}^2-\tilde{\gamma}$, then there exists $\alpha^*<p-1$ such that $\hat G=G$  with probability greater than 
\begin{equation*}
    \begin{aligned}
         &1-4B(K)\alpha^*\exp\left\{-\frac{2}{LK^2\sqrt{M_K}}\left(\delta'\sqrt{n}\right)^{\frac{1}{K}}\right\}\\
         &-\frac{3p(p-1)}{2}\exp\left\{-\frac{1}{2L\sqrt{M_2}}\left(\frac{\lambda\Tilde{\gamma}\sqrt{n}}{2+\lambda}\right)^{\frac{1}{2}}\right\},
    \end{aligned}
\end{equation*}
for all 
$
        n>    \max\left\{{\tfrac{e^2(2+\lambda)^2(4L^2\sqrt{M_2})^4}{\lambda^2\tilde{\gamma}^2}, \tfrac{e^2(LK^2\sqrt{M_K})^{2K}}{\delta^{'^2}}}\right\}.
$
\end{theorem}

{\bf Computational Complexity.}
The complexity of the three algorithms is dominated by the $\mathcal{O}(p^2log(p))$ cost of the Kruskal algorithm which computes the Chow--Liu tree, see~\cite[Chapter VI]{cormen:introduction}. In terms of the edge orientation, Algorithms~\ref{alg:samp:pair} have linear computational complexity both in $p$ and $n$ which is independent of the structure of the graph, while Algorithms~\ref{alg:samp:trip_pair} and~\ref{alg:samp:pair_trip} may entail a cost that is quadratic in $p$ in the worst case scenario, e.g., a star tree with all the edges outgoing from the center.

\section{Numerical Experiments}
\label{sec:numerical-experiments}

We assess and compare the accuracy of our three proposed algorithms on synthetic data, simulated as follows: For any fixed choice of $n$, $p$, and error distribution, we first generate a random undirected tree with $p$ nodes using randomly generated Pr\"{u}fer sequences \citep{prufer} and then independently orient each edge. Next, we draw $n$ samples for every node from the error distribution and uniformly draw the coefficients $\lambda_{ij}$ from the interval $(-1, -0.3)\cup(0.3, 1)$. Finally, we multiply the matrix of sampled errors by the matrix $(I-\Lambda)^{-1}$ to obtain samples corresponding to the LSEM defined by the generated polytree. 

\begin{figure*}
  \centering
  \subfloat[Gamma Distribution\label{fig:gamma_ld}]{%
  \includegraphics[scale=1,width=0.5\linewidth]{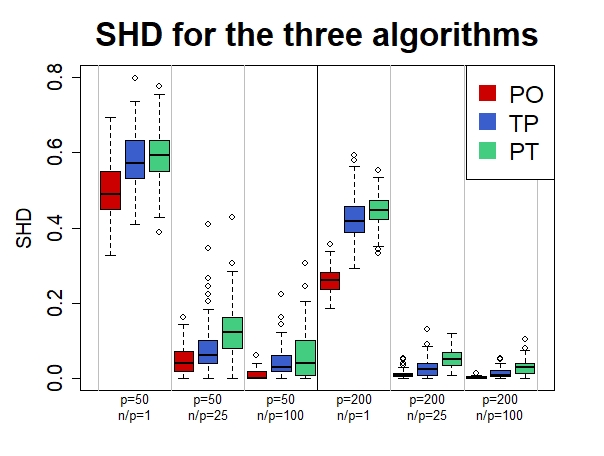}%
  }\hfill
  \subfloat[Uniform Distribution\label{fig:unif_ld}]{%
  \includegraphics[scale=1,width=0.5\linewidth]{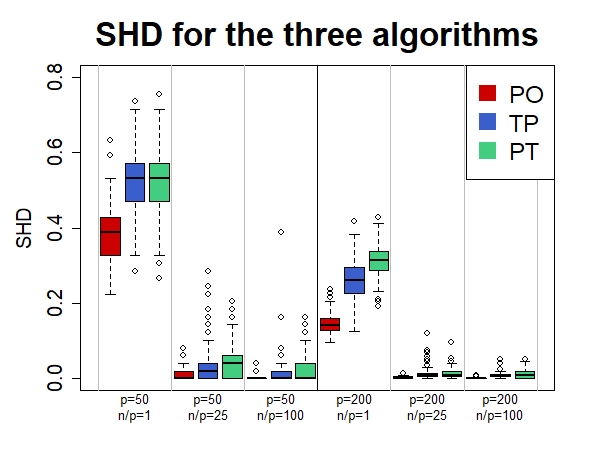}%
  }\\
  
  \subfloat[Gamma Distribution\label{fig:gamma_hd}]{\includegraphics[scale=1,width=0.5\linewidth]{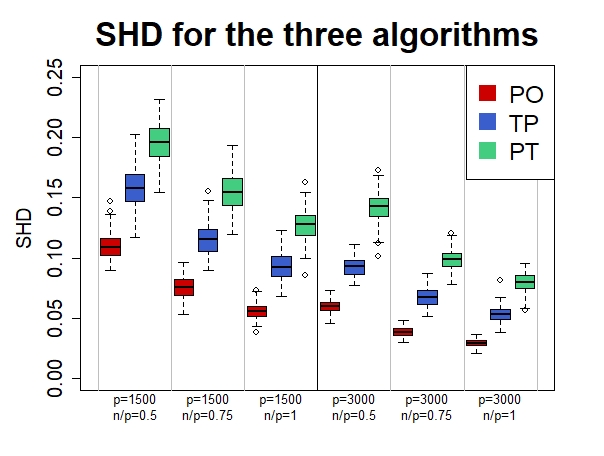}}
  \subfloat[Uniform Distribution\label{fig:unif_hd}]{\includegraphics[scale=1,width=0.5\linewidth]{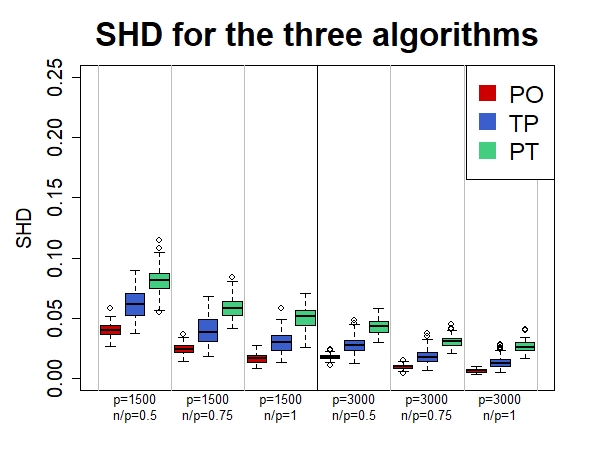}}\\
  
  \subfloat[Gamma Distribution\label{fig:gamma_shd}]{\includegraphics[scale=1,width=0.5\linewidth]{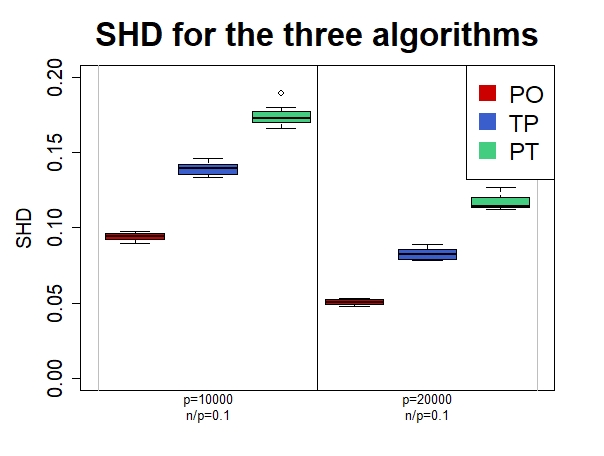}}
  \subfloat[Uniform Distribution\label{fig:unif_shd}]{\includegraphics[scale=1,width=0.5\linewidth]{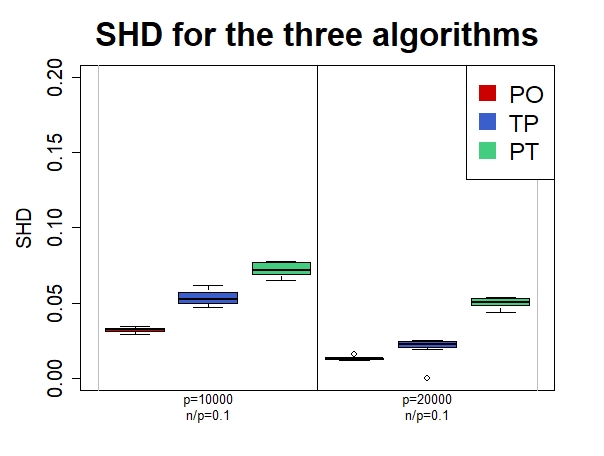}}
  \caption{Performance for low dimensional (\ref{fig:gamma_ld}, \ref{fig:unif_ld}), high dimensional (\ref{fig:gamma_hd}, \ref{fig:unif_hd}) and large scale experiment (\ref{fig:gamma_shd}, \ref{fig:unif_shd}) over 200, 100 and 10 runs, respectively. \label{fig:plots}}
\end{figure*}

The performance is measured by the structural Hamming distance, which is the number of incorrectly included edges plus the number of incorrectly omitted edges, plus the number of incorrectly oriented edges, divided by $2(p-1)$.  Small distance indicates improved performance. We show the results in three settings: (i) low dimensional, with $p\leq 200$ and $1\leq n/p\leq 100$; (ii) high dimensional, with $1500\leq p\leq 3000$ and $0.5\leq n/p\leq 1$; and (iii) a large scale setting with $10000\leq p\leq 20000$ and $n/p=0.1$.  We set up experiments with errors drawn from the gamma and uniform distributions; the results are displayed in Figure \ref{fig:plots}.

For the choice of threshold required in Algorithms~\ref{alg:samp:trip_pair} and~\ref{alg:samp:pair_trip}, we evaluate the algorithms on a grid of thresholds and report the value corresponding to the best result.

{\bf Gamma Distribution.}  Errors were drawn from $\Gamma(\alpha,\beta)$; the shape $\alpha$ and the scale $\beta$ parameters are uniformly drawn from $(0.5,5)$. Since $\Gamma(\alpha,\beta)$ is asymmetric, we tested the algorithms with $K=3$. 

The experimental results are coherent with our developed theory: for all three algorithms, the distance between the true and learned trees converges to 0 as the sample size and/or the dimension of the tree increases.  We observe that 
Algorithm~\ref{alg:samp:pair} performs better both in mean accuracy and variance, despite heavily relying on higher moments which is statistically disadvantageous. The improved performance is potentially due to
the fact that Algorithm~\ref{alg:samp:pair} avoids any potential error propagation since the edges are oriented independently.

{\bf Uniform Distribution.}  Errors were drawn from $U(a,b)$, with the parameter $a$ uniformly drawn from $(-10,-1)$ and $b$ uniformly drawn from $(1,10)$. Here, the uniform distribution is symmetric so the third order cumulants are 0; we thus tested the algorithms with $K=4$. 


The experimental results here are also consistent with our developed theory. We also notice that overall, the experiments with uniform errors outperform those with gamma errors, which may be due to the greater higher order moments associated with the gamma distribution, which tend to increase the variance of the sample cumulants in Corollary~\ref{cor:log:cum}.

The code to reproduce the experiments is available at  \href{https://github.com/danieletramontano/LiNGAM-Polytree-Learning}{https://github.com/danieletramontano/LiNGAM-Polytree-Learning}.
\section{Conclusion}
\label{sec:conclusion}

In this paper, we proposed three algorithms that learn linear non-Gaussian polytree models first using the Chow--Liu algorithm to infer the graph skeleton, and then subsequently applying different approaches to orient edges leveraging non-Gaussianity and marginal uncorrelatedness.  The algorithms differ from one another in how much information is taken from correlations versus from higher moments.  The numerical experiments show that the algorithms also perform well in very high-dimensional problems.  These results indicate that our approach may be applicable in preliminary data analyses towards the aim of understanding dependence structures in data, particularly since the polytree setting allows for richer dependence and causal structures than other tree-based models \citep[e.g.,][]{edwards:2010}.




Our work motivates the following questions for future research:

{\em How to avoid Chow--Liu?} As pointed out above the main computational burden comes from the computation of the Chow--Liu tree. Another shortcoming of the Chow--Liu algorithm is that it requires the whole covariance matrix to be computed and stored beforehand, making it impractical for very large graphs. A solution to this problem that leverages on algebraic relations has been proposed by~\cite{Lugosi:2021} for undirected trees. A possible extension of this approach to polytrees would be of interest.

\begin{figure}
\centering
        \includegraphics[width=1\linewidth]{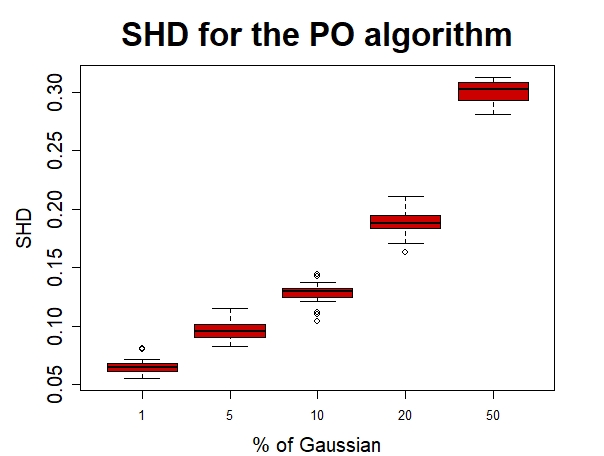}
    \caption{Performance of Algorithm~\ref{alg:samp:pair} for high dimensional experiments with varying \% of Gaussian random variables over 25 runs.}
    \label{fig:gauss}
\end{figure}

{\em How to best handle Gaussian random variables when learning a polytree?}
In some settings we may encounter the situation that some but not all errors are non-Gaussian; see \cite{hoyer:2008} for a characterization of equivalence of graphs in this case.  An interesting problem is then to determine how the respective performance of our algorithms is affected by partial Gaussianity and provide modifications that effectively learn a polytree equivalence class. 
As an illustration, Figure~\ref{fig:gauss} shows that Algorithm~\ref{alg:samp:pair} achieves $70\%$ accuracy when a random choice of half of the random variables are allowed to be Gaussian. 

{\em Which tree structures are the most difficult to learn?}
    \cite{tan:2009} show that for undirected Gaussian tree models, the star and the chain represent the most difficult and the easiest trees to learn, respectively. The difficulty is due to the correlation decay.  
    An interesting question to pursue is what the polytree analogues for the most difficult and easiest trees to learn would be.
    
{\em What happens when the graph is not a tree?}
    \cite{Bresler:2021} prove a weakness result of the Chow--Liu algorithm under model misspecification for the Ising model and adapt it to achieve a form of optimality. It would be of interest to describe a similar optimality criterion in the LiNGAM setting and investigate how our algorithm performs under these terms. 

\begin{acknowledgements} 
This project has received funding from the European Research Council (ERC) under the European Union’s Horizon 2020 research and innovation programme (grant agreement No 883818). DT's PhD scholarship is funded by the IGSSE/TUM-GS via a Technical University of Munich--Imperial College London Joint Academy of Doctoral Studies (JADS) award (2021 cohort, PIs Drton/Monod).
\end{acknowledgements}


\appendix
\section{Lemmas and Proofs}
\label{app:proofs}
\subsection{Section 2}
\begin{example}
\label{ex:two:vert}
Assume we have two random variables $X_1$ and $X_2$, coming from an unknown xSEM, and we wish to find the DAG $G$ that generates the model.  Assume further that $X_1$ and $X_2$ are correlated, leaving us with the two options
$G_1=([2],\{1\to 2\})$ and $G_2=([2],\{2\to 1\})$.  Under the model given by $G_1$, we have $X_1=\varepsilon_1$ and $X_2=\lambda_{1 2}X_1+\varepsilon_2$, and the covariance matrix from \eqref{eq:Sigma} becomes
\begin{equation}
    \Sigma=
        \begin{bmatrix}
            \varepsilon^{(2)}_1 & \lambda_{1 2}\varepsilon^{(2)}_1\\
            \lambda_{1 2}\omega_1 & \lambda^2_{1 2}\varepsilon^{(2)}_1+\varepsilon^{(2)}_2
        \end{bmatrix}.
\end{equation}
Observe that any positive definite $2\times 2$ matrix $\Sigma$ can be written in this way;  set $\varepsilon^{(2)}_1=\Sigma_{1 1}>0$, $\lambda_{1 2}=\Sigma_{2 1}/\Sigma_{1 1}$ and $\varepsilon^{(2)}_2=\Sigma_{2 2}-\Sigma_{1 2}^2/\Sigma_{1 1}>0$.  By symmetry, the model given by $G_2$ also allows its covariance matrices to be any positive definite $2\times 2$ matrix.  Hence, the sets of covariance matrices resulting from $G_1$ versus $G_2$ are the same, and nothing can be said about the graph on the basis of covariances alone.

Now we see how to solve this identifiability issue using higher order cumulants. The multi-trek rule yields that
\begin{equation*}
    \begin{aligned}
            \mathcal{C}^{(3)}_{111}&=\varepsilon^{(3)}_1, & \mathcal{C}^{(3)}_{112}&=\varepsilon^{(3)}_1\lambda_{12},\\
                \mathcal{C}^{(3)}_{122}&=\varepsilon^{(3)}_1\lambda_{12}^2,
            &\mathcal{C}^{(3)}_{222}&=\varepsilon^{(3)}_1\lambda_{12}^3+\varepsilon^{(3)}_2.
    \end{aligned}
\end{equation*} 
We observe a set of simple relations that imply that the matrix 
\begin{equation*}
    A^{1\to 2,3}=
        \begin{bmatrix}
            \Sigma_{11} & \mathcal{C}^{(3)}_{111} & \mathcal{C}^{(3)}_{112}\\
            \Sigma_{12} & \mathcal{C}^{(3)}_{112} & \mathcal{C}^{(3)}_{122}
        \end{bmatrix}
\end{equation*}
has rank 1; see also \cite{wang:drton:2020} where the first two columns of the matrix are considered.  Indeed, the second row of the matrix equals $\lambda_{12}$ times the first row.  
However, this rank constraint generally does not hold in cumulants $(\tilde\Sigma,\tilde{\mathcal{C}}^{(3)})\in\mathcal{M}^{(\le 3)}(G_2)$, where $G_2$ is the DAG with edge $1\xleftarrow{}2$.  A straightforward calculation confirms that the rank of $A^{1\to 2,3}(\tilde\Sigma,\tilde{\mathcal{C}}^{(3)})$ drops to one iff $\lambda_{21}=0$ or $\varepsilon^{(3)}_1=\varepsilon^{(3)}_2=0$.  In other words, for correlated variables $X_1$ and $X_2$ generated with at least one of the errors non-Gaussian with nonzero third moment, the rank constraint on $A^{1\to 2,3}$ discriminates $G_1$ and $G_2$.
these relations do not hold, in general, when the model is generated from the other graph, so checking if these relations holds or not give us a way to identify the right graph, up to a measure 0 set of parameters given by the intersection of $\mathcal{M}^{(\leq3)}(G_1)$ and $\mathcal{M}^{(\leq3)}(G_2)$. 
\end{example}

\begin{proof}[Proof of Corollary~\ref{cor:simple-trek-rule}]
    We say that a $k$-trek $T=(P_1,\dots,P_k)$ factorizes through $k$-trek $S=(Q_1,\dots,Q_k)$ if $Q_j\subset P_j$ for all $j$.   Indeed, it is easy to see that $\lambda^T=\lambda^{S}\lambda^{T-S}$, where  $T-S=(P_1-Q_1,\dots,P_k-Q_k)\in\mathcal{T}^S=\mathcal{T}(\ttop(S),\dots,\ttop(S))$ with $P_j-Q_j$ being the directed path from $\ttop(T)$ to $\ttop(S)$ that remains after removing the edges in $Q_j$ from $P_j$.

For the sake of readability, when the considered set of vertices is clear from the context, we denote the set of treks (and simple treks) by only $\mathcal{T}$ (and $\mathcal{S})$.  Now note that every trek factorizes along one and only one simple trek.  Hence, we can partition $\mathcal{T}=\dot{\bigcup}_{S\in \mathcal{S}}\mathcal{T}^{S}$, according to the simple trek through which the factorization occurs.  The expression in \eqref{eq:trek} may thus be rewritten as
\begin{equation*}
\begin{aligned}
        \mathcal{C}^{(k)}_{i_1,\dots,i_k}(G)=\displaystyle\sum_{S\in \mathcal{S}}\lambda^{S}\left(\displaystyle\sum_{T\in\mathcal{T}^{S}}\varepsilon^{(k)}_{\ttop(T)}\lambda^{T-S}\right).
\end{aligned}
\end{equation*}
By the multi-trek rule, the term in parentheses is  $\mathcal{C}^{(k)}_{\ttop(S)}(G)$.
\end{proof}

\begin{proof}[Proof of Corollary~\ref{cor:simple_trek_2}]
Since the graph is acyclic, the only trek in $\mathcal{T}(i,..,i)$ that has $i$ as $top$ is the trivial trek, from which come the $\varepsilon_i^{(k)}$ term. All the other treks in $\mathcal{T}(i,..,i)$ factorize through a set of distinct parents of $i$, so we can write the sum in~\ref{eq:trek} as
\begin{equation*}
    \displaystyle\sum_{p_1,\dots,p_k\in \pa(i)}\lambda_{p_1, i}\cdots\lambda_{p_k,i}(\displaystyle\sum_{T\in\mathcal{T}(p_1,..,p_k)}\lambda^T\varepsilon^{(k)}_{top(T)})+\varepsilon_i^{(k)}
\end{equation*}
and the term inside the internal parenthesis is $\mathcal{C}^{(k)}_{p_1,\dots,p_k}(G)$.

\end{proof}

\begin{proof}[Proof of Proposition~\ref{prop:rank}]
  By the simple multi-trek rule for polytrees 
$
  c^{e,k}_m =\lambda_{ij}^{k-m}\mathcal{C}_{i}^{(k)}.
$
  Therefore, $c^{e,k}_m=\lambda_{ij}c^{e,k}_{m+1}$ so that the second row of $A^{e,K}$ equals $\lambda_{ij}$ times the first row.
\end{proof}

\subsection{Section 3}
\begin{proof}[Proof of Proposition~\ref{prop:cl:corr}]
As noted, $\mathcal{M}(R)$ may be computed using Kruskal's algorithm, which considers all edges in decreasing order of their weights and adds them to the spanning tree if their presence does not create a (undirected) cycle.

Let $\mathcal{S}(G)$ be the skeleton of $G$.  For a contradiction, assume that $\mathcal{M}(R)\not=\mathcal{S}(G)$.  Since both graphs are trees, we have $L:=\mathcal{M}(R)\setminus\mathcal{S}(G)\neq\emptyset$. Take $\tilde{e}=\{i,j\}=\argmax_{e\in L}|\rho_{e}|$.  Then $|\rho_{\tilde{e}}|\neq0$ and, thus, $\mathcal{T}(i,j)\neq\emptyset$.   The unique trek in $G$ that connects $i$ and $j$ must contain an edge $e$ that is not in $\mathcal{M}(R)$; otherwise we would have two paths between $i$ and $j$ in $\mathcal{M}(R)$ which cannot occur as $\mathcal{M}(R)$ is a tree.  Moreover, Wright's formula from Lemma~\ref{lem:wright} and the assumption made on the correlation coefficients imply that $|\rho_{\tilde{e}}|<|\rho_{e}|$.  But then $e$ would appear before $\tilde{e}$ in Kruskal's algorithm and $e$ would be added to $\mathcal{M}(R)$ since all the edges in $\mathcal{M}(R)$ with a weight higher than the weight of $\tilde{e}$ are correctly classified and so the presence of $e$ could not create a loop. We have arrived at a contradiction.
\end{proof}

\begin{proof}[Proof of Theorem~\ref{theo:generic_cum}]
The first claim is a restatement of Proposition~\ref{prop:rank}. To prove the second claim, we need to show that for generic error cumulants, at least one of the $2\times 2$ subdeterminants of $A^{j\to i,K}$ is nonzero. Since these minors are polynomials in the cumulants, it is enough to prove that they are not identically zero; see, e.g., the lemma in \cite{okamoto:1973}.

By the simple multi-trek rule (Corollary~\ref{cor:simple-trek-rule}),
\begin{align*}
    &c^{j\to i,2}_i=\Sigma_{i,j}=\lambda_{i,j}\Sigma_{i,i}, \\
    &c^{j\to i,k}_m=\lambda^m_{i,j}\mathcal{C}^{(k)}_i \quad\forall m<k\leq K.
\end{align*}
By Corollary~\ref{cor:simple_trek_2},
\begin{align*}
    c^{j\to i,2}_2=\Sigma_{j,j}&=\sum_{p,q\in\pa(i)}\lambda_{p,j}\lambda_{q,j}\Sigma_{p,q}+\varepsilon^{(2)}_j\\
    &=\sum_{p\in\pa(j)}\lambda_{p,j}^2\Sigma_{p,p}+\varepsilon_j^{(2)},
\end{align*}
because in a polytree any two distinct parents $p$ and $q$ have $\mathcal{T}(p,q)=\emptyset$ and, thus, $\Sigma_{p,q}=0$ by Lemma~\ref{lem:wright}.  

Consider now a minor involving the first column together with any other column with $m<k$.  This minor is
\begin{equation*}
    \begin{aligned}
        &\Sigma_{j,j}c^{j\to i,k}_{m-1}-\Sigma_{i,j}c^{j\to i,k}_{m}\\
        &=\Big[\sum_{p\in\pa(j)}\lambda_{p,j}^2\Sigma_{p,p}+\varepsilon_j^{(2)}\Big]\lambda^{m-1}_{i,j}\mathcal{C}^{(k)}_i-\lambda_{i,j}\Sigma_{i,i}\lambda^{m}_{i,j}\mathcal{C}^{(k)}_i\\
        &=\lambda^{m-1}_{i,j}\mathcal{C}^{(k)}_i\Big[\sum_{p\in\pa(j)}\lambda_{p,j}^2\Sigma_{p,p}+\varepsilon_j^{(2)}-\lambda^2_{i,j}\Sigma_{i,i}\Big]\\
        &=\lambda^{m-1}_{i,j}\mathcal{C}^{(k)}_i\Big[\sum_{p\in\pa(j)\setminus{i}}\lambda_{p,j}^2\Sigma_{p,p}+\varepsilon_j^{(2)}\Big].
    \end{aligned}
\end{equation*}
The term in parentheses is always positive, while the front factor is nonzero provided $\lambda_{i,j}\not=0$ and the error cumulants are chosen such that $\mathcal{C}^{(k)}_i\not=0$ (e.g., take any distribution with $\varepsilon_k^{(k)}=0$, for every $k\neq j$, and $\varepsilon^{(k)}_j\neq0$).
\end{proof}

\begin{proof}[Proof of Proposition~\ref{prop:MEC}]
The claim follows from Lemma~\ref{lem:wright} as $i\to j\xleftarrow{}l$ is the only case in which there are no treks between $i,l$.
\end{proof}

\begin{proof}
From Theorem~\ref{theo:generic_cum} we derive the correctness of the rank condition and thus of the entire algorithm  \textit{PairwiseOrientation\_Pop} as well as the relevant parts of the other two algorithms.  Proposition~\ref{prop:MEC} yields the correctness of the remaining parts of the other two algorithms. 
\end{proof}

\subsection{Section 4}

\begin{lemma}
\label{lemma:conc:ineq}
Consider a degree $k$ polynomial $f(X)=f(X_1,\dots,X_m)$, where $X_1,\dots,X_m$ are possibly dependent random variables with log-concave joint distribution on $\mathbb{R}^m$. Then exists a constant $L>0$ such that for all $\delta$ with
\begin{equation*}
    \frac{2}{L}\left(\frac{\delta}{e\sqrt{\var[f(X)]}}\right)^{\frac{1}{K}}>2,
\end{equation*}
we have
\begin{equation*}
    \begin{aligned}
            &\mathbb{P}[|f(X)-\mathbb{E}[f(X)]|>\delta] \\    &\leq\exp\left\{-\frac{2}{L}\left(\frac{\delta}{\sqrt{\var[f(X)]}}\right)^{\frac{1}{K}}\right\}.
    \end{aligned}
\end{equation*}
\end{lemma}
\begin{proof}[Proof of Corollary ~\ref{cor:log:cum}]
The results follows by bounding the variance of $\hat{c}^{(i,j),k}_m$.  To this end, we may express the variance as a linear combination of products of moments based on the definition of cumulants.  We may then bound each product of moments by a power of $M_K$ and note that no weight may exceed $(K-1)!$.  Applying Stirling's approximation for the factorial and bounding the Bell numbers that count the number of summands gives the result.
\end{proof}

\begin{proof}[Proof of Theorem~\ref{theo:chow:cons}]
Let $F$ be the event defined in Lemma~\ref{lemma:lower:bound}.  Then 
\begin{equation*}
    \begin{aligned}
    &\mathbb{P}(\mathcal{M}_n(G)=\mathcal{S}(G))\geq\mathbb{P}(F)\\
    &\geq 1-\sum_{i\le j} \mathbb{P}(|\hat{\rho}_{i,j}-\rho_{i,j}|>\gamma),
    \end{aligned}
\end{equation*}
where the last inequality comes from the union bound. 

Correlations are scale-invariant, thus the value of $\rho_{i,j}$ and the distribution of $\hat\rho_{i,j}$ do not change under rescaling of the observed variables to $\tilde{X}_i=X_i/\lambda$.  Now, we apply Lemma~\ref{lemma:sym:matrices} with $A=\Tilde{\Sigma}$ and $B=\Tilde{\Sigma}^*$ as the new population and sample covariance matrices, respectively. Let $||\Tilde{\epsilon}||_{\infty}=||\Tilde{\Sigma}- \Tilde{\Sigma}^*||_{\infty}$. Then $||\Tilde{\epsilon}||_{\infty}\leq\frac{||\epsilon^{(i,j),2}||_{\infty}}{\lambda}$, where we indicate with $\epsilon^{(i,j),2}$ the vector containing the errors $\epsilon^{(i,j),2}_m$, for $m=0,1,2$.  This allows us to write
\begin{equation*}
    \begin{aligned}
        \mathbb{P}(|\rho^*_{i,j}-\rho_{i,j}|>\gamma)&\leq\mathbb{P}(||\Tilde{\epsilon}||_\infty>\frac{\gamma}{2+\gamma})\\
        &\leq\mathbb{P}(||\epsilon^{(i,j),2}||_\infty>\frac{\lambda\gamma}{2+\gamma})\\ &\leq3\exp\left\{-\frac{1}{2L\sqrt{M_2}}\left(\frac{\lambda\gamma\sqrt{n}}{2+\lambda}\right)^{\frac{1}{2}}\right\},
    \end{aligned}
\end{equation*}
where the last inequality comes from a union bound and Corollary~\ref{cor:log:cum} with $K=2$.
\end{proof}

\begin{proof}[Proof of Lemma~\ref{lem:taylor}]
\label{Taylor}
By Taylor expansion, we have
\begin{equation}
    \begin{aligned}
    f(c+\epsilon)-f(c)=\nabla(f)_{|c}\cdot\epsilon+\frac{\epsilon^t\cdot H(f)_{|c}\cdot\epsilon}{2},
    \end{aligned}
\end{equation}
where $H(f)_{|c}$ is the Hessian matrix of $f$ computed in $c$.  For the special quadratic polynomial $f$, the Hessian is constant with entries $\pm 1$, and the gradient contains entries of $c$, possibly negated.  The result follows by triangle inequality.
\end{proof}

\begin{proof}[Proof of Theorem~\ref{theo:chow:cons}]
The event of correctly reconstructing the polytree $G$ is the intersection of the two events of correctly recovering of the skeleton and correctly orienting all the edges.  The probability of this intersection is bounded from below by the sum of the probabilities of the two events minus one.  Theorem~\ref{theo:chow:cons} and  Lemma~\ref{lem:po:lower:bound} imply the result.
\end{proof}

\begin{proof}[Proof of Lemma~\ref{lem:po:lower:bound}]
Given that we start with the correct skeleton, a mistake in the orientation appears if and only if there is an edge, $e$ in the skeleton for which $||\hat{v}_{r}(e)||\geq||\hat{v}_{w}(e)|$, so the union bound leads to the following lower bound 
\begin{equation*}
    \begin{aligned}
        &\mathbb{P}(\mathcal{A}_n^{PO}(E_{\mathcal{S}(G)},K)=G)\\
        &=\mathbb{P}(||\hat{v}_{r}(e)||<||\hat{v}_{w}(e)||, \forall e\in\mathcal{S}(G))\\
        &\geq1-(p-1)\mathbb{P}(||\hat{v}_{r}(e)||\geq||\hat{v}_{w}(e)||).
    \end{aligned}
\end{equation*}
Now we need an upper bound on $\mathbb{P}(||\hat{v}_{r}(e)||\geq||\hat{v}_{w}(e)||)$, adding and subtracting $||v_{w}(e)||$ on the right hand side of the inequality, and using $||v_{w}(e)||>\delta$, led to
\begin{equation*}
    \begin{aligned}
    &\mathbb{P}(||\hat{v}_{r}(e)||\geq||\hat{v}_{w}(e)||)\\
    &\leq\mathbb{P}(||\hat{v}_{r}(e)||+(||v_{w}(e)||-||\hat{v}_w(e)||)\geq\delta)\\
    &\leq\mathbb{P}(||\hat{v}_{r}(e)||+(||v_{w}(e)-\hat{v}_w(e)||)\geq\delta)\\
    &\leq\mathbb{P}(||\hat{v}_{r}(e)||\geq\tfrac{\delta}{2})+\mathbb{P}(||v_{w}(e)-\hat{v}_w(e)||\geq\tfrac{\delta}{2})\\
    &=\mathbb{P}(||\hat{v}_r(e)||^2\geq\tfrac{\delta^2}{4})+\mathbb{P}(||v_{w}(e)-\hat{v}_w(e)||^2\geq\tfrac{\delta^2}{4}).
    \end{aligned}
\end{equation*}
Now, using that $v_r(e)=0$ and applying Lemma~\ref{lem:taylor}, we obtain that
\begin{equation*}
    \begin{aligned}
    &\mathbb{P}(||\hat{v}_r(e)||^2\geq\tfrac{\delta^2}{4})+\mathbb{P}(||v_{w}(e)-\hat{v}_w(e)||^2\geq\tfrac{\delta^2}{4})\\
    &\leq\mathbb{P}(||\hat{v}^i_r(e)||^2\geq\tfrac{\delta^2}{4B(K)})+\\
    &\qquad+\mathbb{P}(||v^i_{w}(e)-\hat{v}^i_w(e)||^2\geq\tfrac{\delta^2}{4B(K)})\\
    &\leq2\mathbb{P}(4M_K||\epsilon||_{\infty}+2||\epsilon||_{\infty}^2\geq\tfrac{\delta}{2\sqrt{B(K)}}),
    \end{aligned}
\end{equation*}
where $v^i$ is one of the components of the vector. Finally, we obtain that
\begin{equation*}
    \begin{aligned}
        &\mathbb{P}(4M_K||\epsilon||_{\infty}+2||\epsilon||_{\infty}^2\geq\tfrac{\delta}{2\sqrt{B(K)}})\\
        &\leq\mathbb{P}(||\epsilon||_{\infty}\geq\tfrac{\delta}{4M_K\sqrt{B(K)}})+\mathbb{P}(||\epsilon||_{\infty}\geq\tfrac{\sqrt{\delta}}{\sqrt[4]{4B(K)}}).
    \end{aligned}
\end{equation*}
With $\delta':=\min\{\frac{\delta}{4M_K\sqrt{B(K)}},\frac{\sqrt{\delta}}{\sqrt[4]{4B(K)}}\}$ we can bound the right hand side of the last inequality by
\begin{equation*}
    \begin{aligned}
        2\mathbb{P}(||\epsilon||_\infty\geq\delta')\leq 2B(K)\exp\left\{-\tfrac{2}{LK^2\sqrt{M_K}}\left(\delta'\sqrt{n}\right)^{\frac{1}{K}}\right\}
    \end{aligned}
\end{equation*}
whenever $\frac{2}{LK^2\sqrt{M_K}}\left(\frac{\delta'\sqrt{n}}{e}\right)^{\frac{1}{K}}>2$ holds, and this concludes the proof.
\end{proof}

\begin{proof}[Proof of Lemma~\ref{lem:rho_crit}]
It suffices to show that under the event $\tilde{F}=\bigcap\{|\hat{\rho}_{i,j}-\rho_{i,j}|<\tilde{\gamma}\}$ all independence tests give the correct decision.  This claim was proven in Theorem 3.5 of \citep{lou:2021}.
The rest of the proof is the same as for Theorem~\ref{theo:chow:cons}.
\end{proof}

\begin{proof}[Proof of Theorem~\ref{theo:pt_tp:cons}]
Having $\tilde{\gamma}\leq\gamma$ from Lemma~\ref{lemma:lower:bound} and Lemma~\ref{lem:rho_crit}, the probability that the skeleton is recovered correctly and that all the independence tests give the right answer is bounded from below by the probability of the event considered in Lemma~\ref{lemma:lower:bound}. Hence, it only remains to bound the probability that all cumulant tests correctly orient their respective edge. This can be done as in Lemma~\ref{lem:po:lower:bound}. Finally, carrying over the arguments from the proof of Theorem~\ref{theo:po:cons} gives the result. Notice that in the case of Algorithm~\ref{alg:samp:trip_pair}, the constant $\alpha^*<p-1-\textit{CP}$, where $\textit{CP}$ is the number of directed edges in the CPAG associated to $G$. 
\end{proof}

\section{Sample Version of the Algorithms}
\label{app:sec:samp:alg}

\begin{algorithm}[H]\caption{PairwiseOrientation$(E,K)$}
\label{alg:samp:pair}
    \begin{algorithmic}[1]
        \State{$O\gets\emptyset$}
        \For{$e=\{i,j\}\in E$} 
            \If{$||\hat{v}_{i\to j}(\{i,j\})||<||\hat{v}_{j\to i}(\{i,j\})||$}
                \State{$O\gets O\cup\{i\to j\}$}
                \Else\,  \State{$O\gets O\cup\{j\to i\}$}
            \EndIf
        \EndFor
    \Return $O$
    \end{algorithmic}
\end{algorithm}

\begin{algorithm}[H]\caption{PTO$(E,K,\rho_{\theta})$}
\label{alg:samp:pair_trip}
    \begin{algorithmic}[1]
        \State{$O\gets\emptyset$}
        \For{$i-j-k\in E$}
            \If{$|\hat{\rho}_{i,k}|<\rho_{\theta}$}
                \State{$E\gets E\setminus\{\{i,j\},\{j,k\}\}$}
                \State{$O\gets O\cup\{i\to j, k\to j\}$}
            \EndIf
        \EndFor
        \For{$i\to j\in O$}
            \For{$j-l\in E$}
                \State{$E\gets E\setminus\{j-l\}$}
                \State{$O\gets O\setminus\{j\to l\}$}
            \EndFor
        \EndFor
        \For{$\{i,j\}\in E$}
            \If{$||\hat{v}_{i\to j}(\{i,j\})||<||\hat{v}_{j\to i}(\{i,j\})||$}
                \State{$O\gets O\cup\{i\to j\}$}
                \For{$j-l\in E$}
                    \State{$E\gets E\setminus\{j-l\}$}
                    \State{$O\gets O\setminus\{j\to l\}$}
                \EndFor
                \Else 
                    \State{$O\gets O\cup\{j\to i\}$}
                    \For{$i-l\in E$}
                        \State{$E\gets E\setminus\{i-l\}$}
                        \State{$O\gets O\setminus\{i\to l\}$}
                    \EndFor
            \EndIf
        \EndFor
        \Return{O}
    \end{algorithmic}
\end{algorithm}
\vfill\mbox{ }

\begin{algorithm}[H]\caption{TPO$(E,K,O,o,\rho_{\theta})$}
\label{alg:samp:trip_pair}
    \begin{algorithmic}[1]
        \If{$E\neq\emptyset$}
            \If{$o=\emptyset$}
                \State{$\{i,j\}\gets E[1]$}
                \If{$||\hat{v}_{i\to j}(\{i,j\})||<||\hat{v}_{j\to i}(\{i,j\})||$}
                    \State{$o\gets (i\to j)$}
                    \State{$O\gets O\cup\{o\}$}
                    \Else  
                        \State{$o\gets (j\to i$)}
                        \State{$O\gets O\cup\{o\}$}
                \EndIf
            \EndIf
            \State{$E_o\gets E\cap t(o)$}
            \If{$E_o\neq\emptyset$}
                \State{$E\gets E\setminus E_o$}
                \For{$t(o)-k\in E_o$}
                    \If{$|\hat{\rho}_{s(o),k|}<\rho_{\theta}$}
                        \State{$O\gets O\cup\{k\to t(o)\}$}
                        \Else                                                       \State{$O\gets O\cup\{t(o)\to w\}$}
                            \State{$o\gets(t(o)\to w)$}                        
                            \State{$O,E\gets TPO\_Pop(E,K,O,o,\rho_{\theta})$}
                    \EndIf
                \EndFor
            \EndIf
            \State{$O,E\gets TPO\_Pop(E,K,O,\emptyset,\rho_{\theta})$}
        \EndIf
        \Return{O,E}
    \end{algorithmic}
\end{algorithm}

\bibliography{tramontano_678}
\end{document}